# Integration of 4D BIM and Robot Task Planning: Creation and Flow of Construction-Related Information for Action-Level Simulation of Indoor Wall Frame Installation

[1] Hafiz Oyediran, [2] William Turner, [3] Kyungki Kim, [4] Matthew Barrows

[1]Graduate Research Assistant, Durham School of Architectural Engineering and Construction, University of Nebraska- Lincoln, USA; email: hoyediran2@huskers.unl.edu
[2] Graduate Research Assistant, The School of Computing, University of Nebraska - Lincoln, USA; email: william.turner@huskers.unl.edu
[3]Assistant Professor, Durham School of Architectural Engineering and Construction, University of Nebraska-Lincoln; PKI 206D, 1110 S. 67th Street, Omaha, NE 68182-0816, USA; Tel: +01-770-361-2262; email: kkim13@unl.edu (Corresponding Author)
[4]Assistant Professor of Practice, Durham School of Architectural Engineering and Construction, University of Nebraska-Lincoln, USA; email: mbarrows@unl.edu

## Abstract

An obstacle toward construction robotization is the lack of methods to plan robot operations within the entire construction planning process. Despite the strength in modeling construction site conditions, 4D BIM technologies cannot perform construction robot task planning considering the contexts of given work environments. To address this limitation, this study presents a framework that integrates 4D BIM and robot task planning, presents an information flow for the integration, and performs high-level robot task planning and detailed simulation. The framework uniquely incorporates a construction robot knowledge base that derives robot-related modeling requirements to augment a 4D BIM model. Then, the 4D BIM model is converted into a robot simulation world where a robot performs a sequence of actions retrieving construction-related information. A case study focusing on the interior wall frame installation demonstrates the potential of systematic integration in achieving context-aware robot task planning and simulation in construction environments.

Keywords: Building information building (BIM); 4D BIM; construction planning; system integration; information flow; knowledge base; construction robotics; robot task planning; robot operating system (ROS); robot simulation

Highlights

- Integrated 4D BIM and robotics to simulate robot tasks during construction planning
- Presented a framework for high-level robot task planning and detail simulation
- Presented a construction robot knowledge base for systematic information flow
- Automatically derived additional 4D BIM modeling requirements for robot simulation
- Simulated a mobile robot’s actions to install wall frames in a residential building

## 1. Introduction

Rapid advancements in robotics technologies are making the utilization of robots for dangerous, tedious, and repetitive tasks more and more practical [1]. Unlike traditional industrial robots with fixed behaviors, modern robots with mobile platforms, sensors, and actuators can be programmed to perform given tasks intelligently adapting to changing work environments. Many sectors, including manufacturing [2], rescue [3], agriculture [4], and healthcare [5], are adopting robots to automate existing processes to achieve greater productivity and safety. According to [6], currently available Artificial Intelligence (AI) and robotics technologies can automate a wide range of tasks in major US industry sectors, such as accommodation and food services (74% automation potential), manufacturing (60%), agriculture (58%), transportation and warehousing (57%), retail trade (53%), mining (51%), and construction (47%). Many construction tasks are repetitive and labor-intensive by nature [7,8], and thus robotization of these tasks can potentially address many chronic problems, such as stagnant productivity growth [9], labor shortage [10], and work-related diseases/fatalities [11]. A growing number of robotic solutions are introduced by academic studies [12,13] and industrial applications (excavation and leveling [14], marking of layout [15], rebar tying [16], and bricklaying [17,18]). With this trend, construction sites are expected to become crowded with robots and human workers in the near future exposing human workers to robot-related hazards, such as collisions, crushing, trapping, mechanical part accidents, etc. [19]. In order to utilize robots safely and effectively in congested construction environments, both high-level task planning and detailed simulation of construction robots should be performed as part of the entire construction planning. Despite the abundant studies on the coordination between human work crews [20,21], none of the prior studies incorporated robot operations into construction planning process.

The long-term goal of this study is to create construction planning technology that can generate plans for construction robots and analyze their coordination with human workers within the construction environment. Throughout this paper, robot task planning refers to efforts involved in both high-level task planning and detailed simulation of construction robots. A robot can complete a task (e.g., painting, frame installation, material delivery) by performing a sequence of actions interacting with its work environment and the user [22]. High-level robot task planning for a certain task (e.g., frame installation) involves identifying all the required actions (e.g., navigate to material storage, pick material, navigate to install location, place material), determining the sequence between the actions, and matching these individual actions with robot skills (e.g., navigate, pick, place) that are the robot's functionalities to perform the corresponding actions. On the other hand, detailed task simulation performed after the high-level task planning executes the robot skills generating precise motions considering the given spatiotemporal condition of the work environment. Recent studies in the development of intelligent robot behaviors in built environments utilize contextual information from the work environments either via real-time sensor-based perception or as embedded knowledge as the input to execute robot skills [23,24]. For example, a navigation skill of a robot can be triggered by providing the coordinate of a material storage as the input argument. Given the coordinates of the destination and a map of the environment, the navigation skill generates a navigation path and travels to the material storage while updating the navigation path with observed obstacles. Speed limits can be specified for different regions on the map or directly determined by a user.

A significant challenge in performing robot task planning in construction is the lack of construction planning techniques that can define robot tasks and realistically simulate the required actions within the construction environment. Currently, 4D Building Information Modeling (4D BIM) is the most advanced technology to simulate changing construction site conditions [25]. A 4D BIM model visualizes a construction process by making 3D building elements (e.g., walls, columns, floors) and non-building elements (e.g., temporary structures, workspaces, equipment) appear and disappear according to a construction schedule. Several studies have demonstrated that 4D BIM can be used to simulate workspace plans [26–29], heavy equipment operations [30,31], waste recycling and reuse plans [32], and site logistics [33,34]. However, despite the strength in describing construction site conditions, existing 4D BIM technologies do not support the high-level task planning as well as detailed simulation that is uniquely necessary for construction robot operations. Also, the currently available robot task planning approaches [35–37] cannot utilize construction-related information in a 4D BIM model due to the systematic separation between the two technologies. By systematically integrating robot task planning with the 4D BIM, a robot's task plan can be generated directly utilizing the contextual information about the work environment available in a 4D BIM model. Recent works that attempted to interface BIM and robotics [24,38] failed to address important research questions related to their systematic integration, such as *"how can the two heterogeneous systems (4D BIM and robot task planning) be interfaced in a way that allows robot task simulation as part of the entire construction planning process?"*, *"how should a 4D BIM model be created to contain the construction-related and robot-related information required for robot task planning?"* and "*how should information flow between the two systems to allow a robot operating software to retrieve contextual information about the construction environment required to perform task planning?"*. Answering these questions, this study presents a framework that connects 4D BIM and robot task planning through the creation of a construction robot knowledge base. As the key connector between 4D BIM and robotics, the construction robot knowledge base binds tasks in a 4D BIM model, robot-oriented definitions of construction tasks, and contextual information of the construction environment as the input to perform the tasks. This study presents how we can create a 4D BIM model containing all the information required for robot task planning, how the information requirements are derived from the construction knowledge base, how a robot simulation environment called robot world is generated from the 4D BIM model, and how a robot simulation is performed in the robot world retrieving contextual information about the work environment.

## 2. Literature Review

This section first reviews studies in the utilization of 4D BIM technologies for various aspects of construction planning introduces robots adopted to perform construction tasks and reviews the state-of-the-art studies that leveraged BIM technologies for construction robot task planning and simulation.

### 2.1. 4D BIM for construction planning

Numerous studies have employed 4D BIM technologies to enhance various areas of construction planning, such as construction scheduling, workspace allocation, equipment and temporary structure planning, etc. Sheikhkhoshkar et al [39] created a 4D BIM-based technology that generates a concrete pouring plan describing optimal pouring start points, pouring directions, and concrete pour volume, that results in joint positions with the least impact on the structural strength. Wang et al [40] detected obstructions between the moving paths of workers and mobile equipment from a cell-based representation of the construction site generated from a 4D BIM model. Ji and Leite [41] integrated 4D BIM and rule-based checking to automate the review process of reviewing plans for tower cranes. The proposed rule-based checking approach expedited the process by automatically analyzing the specifications of tower cranes, site-specific conditions presented in the 4D BIM, and construction methods. Tak et al [42] integrated a path planner with 4D BIM to simulate multiple mobile crane lifting operations. Several studies expanded 4D BIM technologies to perform workspace generation, detection of spatiotemporal conflicts between the workspaces, and resolution of spatial conflicts. Jongeling et al [43] estimated potential productivity losses by calculating distances between work crews based on moving work locations and temporary structures manually specified in a 4D BIM model. Kim et al [44] simulated the moving directions of work crews and scaffolds and quantitatively evaluated multiple workflow options in terms of cost, duration, and safety. Mirzaei et al [29] simulated different combinations of workflow patterns of multiple tasks and analyzed spatial conflicts, potential productivity losses, and constructability issues. Some studies utilized 4D BIM as the platform for logistics management [33,45], safety planning [46,47], image-based progress monitoring [48], and environmental planning and management [49]. Even with these efforts, the state-of-the-art technologies in 4D BIM are not addressing the technical requirements to accommodate robot task planning within the construction planning process. Unlike construction planning for human workers with cognitive capabilities, task planning for robots depends on the precise description of tasks to be performed and the generation of highly detailed robot motions (e.g., moving, picking, turning). Most of the studies reviewed above [40,41,43] perform planning for human workers, temporary structures, and equipment mostly in rough details that are not suitable to simulate granular robot operations. Also, robot operations should be planned and simulated acquiring information about dynamically changing construction site conditions from a 4D BIM model. As will be discussed in section 2.2, none of the existing approaches presented a way to create a 4D BIM model that contains all the robot-related information to perform task planning and utilize it to generate a robot operation plan.

## 2.2. Robotization in construction and utilization of BIM for task planning

The use of robots to perform construction tasks has been presented in several studies. Sharif and Gentry [50] presented a robot arm to pick and fold metal sheets according to patterns described in a canopy design. Firth et al [51] presented a prototype painting robot hand replicating the movements and muscles of a human hand gripping different construction tools. Bruun et al [52] used two robot arms that collaboratively assembled a geometrically complex masonry arch without scaffolds. Paraschο [53] implemented an approach similar to Bruun et al [52] using two robot arms to construct a brick arch vault. Sorour et al [54] developed a prototype mobile robot with a manipulator that can scan walls and perform arm motion planning to paint them. Dharwaman et al [55] proposed an approach to integrate robots into existing scaffolding structures for quick deployment of construction robots. Woo et al [56] developed a vehicle-mounted robot for repainting faded road marks. The proposed system detected faded lane marks needing repainting using image processing. Bosscher et al [57] proposed a cable-mounted robot with contour crafting capability that can dramatically expedite concrete construction processes. Even with these efforts in the development of individual construction robots, planning of robot operations from the perspective of construction planning has not been addressed by most of the existing studies. Due to limited spaces and congestion in construction environments, planning robot operations as part of the entire construction planning process, such as 4D BIM modeling, is crucial for effective and safe utilization of robots.

Recent studies presented approaches to bridge the gap between BIM and robotics technologies. Gao et al [58] simulated two robot arms for assembling lightweight structures using information in a BIM model. This study extracted shapes and joint coordinates of individual building elements and the assembly sequence from a BIM model as input for the pick-and-place operations. Chong et al [59] simulated a robot performing the wood frame assembly. Similar to [58], this study obtained geometric information (such as locations and orientations) of the individual frame elements to implement the pick-and-place of the wood frames. Kim et al [38] simulated a mobile painting robot in a robot simulator called Gazebo [60]. This study converted a BIM model into a robot operating system (ROS)-compatible Simulation Description Format (SDF) file to transfer the building information into a physics-based robot simulation. Later, Kim et al [24] created a semantic robot simulation world from a BIM model where a robot can retrieve project information during task planning and execution. This study uniquely modeled a building as a robot in a URDF format [61] that accurately describes the motions of static building elements (e.g., walls and floors) and dynamic building elements (e.g., doors and windows). Momeni et al [62] simulated robot arms for automated fabrication of reinforcement cages. This study used rebar data (e.g., type, order of placement, tying and gripping points) contained in a BIM model to generate detailed installation instructions, such as the placement sequence and tying instructions of each reinforcement bar. Zhu et al [59] performed multi-robot coordination and task allocation for prefabricated housing focusing on steel frame installation. Relying on the numerical simulation of robot locations and building element locations, this study did not simulate precise actions of robot to perform the installation tasks.

## 2.3. Point of departure

Advancements were made in the integration of BIM and robotics by retrieving information about building elements as the input for task planning [58,59] or generating construction environments for robot simulation [24,38]. These studies [24,38,58,62,63] transferred information in 3D BIM models to various robot simulation platforms, such as Gazebo [60], Webots [64], and CoppeliaSim [65]. Extracting information necessary for robot task planning from BIM models and performing task planning in separate robot simulators allowed the utilization of establishments in the robotics area (such as robot representations, implementations of widely used navigation and manipulation algorithms, and visualization and analysis of sensor data) without redeveloping all the features within BIM platforms. Building on these studies, planning for construction robots within the construction planning process requires further efforts to systematically integrate 4D BIM and robot task planning. Beyond the building design information in a 3D BIM model, a 4D BIM contains construction-related information (such as a building under construction, workspaces, temporary structures, material storages, and safety fences) that interact with construction robots and impact robot behaviors. Robot task planning should generate a robot's work plan considering the spatiotemporal conditions presented in the 4D BIM model and analyze how the work environment is impacted by the robot's actions. Also, due to the need for explicit input for robots, additional information necessary for robot task planning should be identified and modeled in a 4D BIM model in advance. The construction-related information in the 4D BIM should then flow into a robot simulation in a form that can be utilized as input to control robot actions. Furthermore, for each task to robotize, a robot-oriented definition should be created to describe required robot actions, the sequence of these actions, and how the actions are simulated by interacting with objects in the construction environment. Most of these essential capabilities required for high-level task planning and simulation of robots have not been incorporated in any of the existing frameworks for creating 4D BIM models or state-of-the-art studies in construction robotics.

## 3. Objective and scope

Building on the works in BIM-robotic integration, this study presents a framework to systematically integrate 4D BIM and robot task planning into a form that allows high-level task planning and detailed simulation of construction robots. This study uniquely presents a construction robot knowledge base to allow the seamless information flow between construction tasks and a 4D BIM model. For tasks selected for robotization, the construction robot knowledge base automatically derives a list of additional robot-related information that should be incorporated into a conventional 4D BIM model. The 4D BIM satisfying the modeling requirements is then converted into a robot simulation world where detailed construction robot simulation is performed. A robot can perform required actions by extracting information about building elements and construction site objects during task planning and simulation. The task to be robotized in this study is limited to interior wall frame installation performed by a mobile robot. However, the generic process developed in this study can be applied to simulate other construction tasks with other robots. The long-term goal of this study is to create a new process and tools that are required to develop a practical robot utilization plan, analyze the robots' behaviors and coordination with human workers and the work environment, and further specify intelligent and safe robot behaviors for given situations. However, the scope of the study presented in this paper is limited to proposing the overall framework to align the two heterogeneous technologies (4D BIM and robot task planning) with the seamless information flow and generate robot task plans. The major accomplishment of this study is the creation of a new 4D BIM-based process that allows the performance of both high-level task planning and detailed simulation of construction robots as part of the entire construction planning. This foundational work in construction robot task planning can open the doors to various future studies, such as quantitative and qualitative analysis of the generated robot task plans, rescheduling and site coordination to safe human-robot collaboration, and sensing and robot controls for safety.

## 4. Framework for robot task planning and simulation in 4D BIM

### 4.1. Overview of the proposed framework

Studies define terms related to robot task planning, such as tasks, actions, and skills, slightly differently. ANSI/RIA R15.06-2012 [66] defines a task program as a series of instructions and functions defining a task of a robot. Rovida et al 2017 [22] describe a task as a sequence of skills, where a skill is a re-occurring action to execute a procedure. According to Lesire et al 2020 [67], robot skills are elementary functions equipped in a robot system to perform a given mission. In our study, partly following definitions in Rovida et al 2017 [22], we distinguish robot actions from robot skills and consider a robot action as a customization of a generic robot skill in an effort to consider the context of the given task. For example, a generic robot navigation skill can be customized for two different actions, navigation carrying a material and navigation without a material. These two actions implement the generic navigation skill with distinct arm positions, moving speeds, and safety stopping distances. In our study, a **robot task** is a sequence of robot actions, a **robot action** is a sub-element of a task which is an implementation of a robot skill considering the context of the work, and a **robot skill** is a generic ability equipped in a robot. A robot skill can be triggered without any input, information acquired from the working environment, or input directly from the user. We also describe a **robot task specification** as a comprehensive robot-oriented description of a task that explains the sequence of actions, and customization of the generic robot skills to execute the actions.

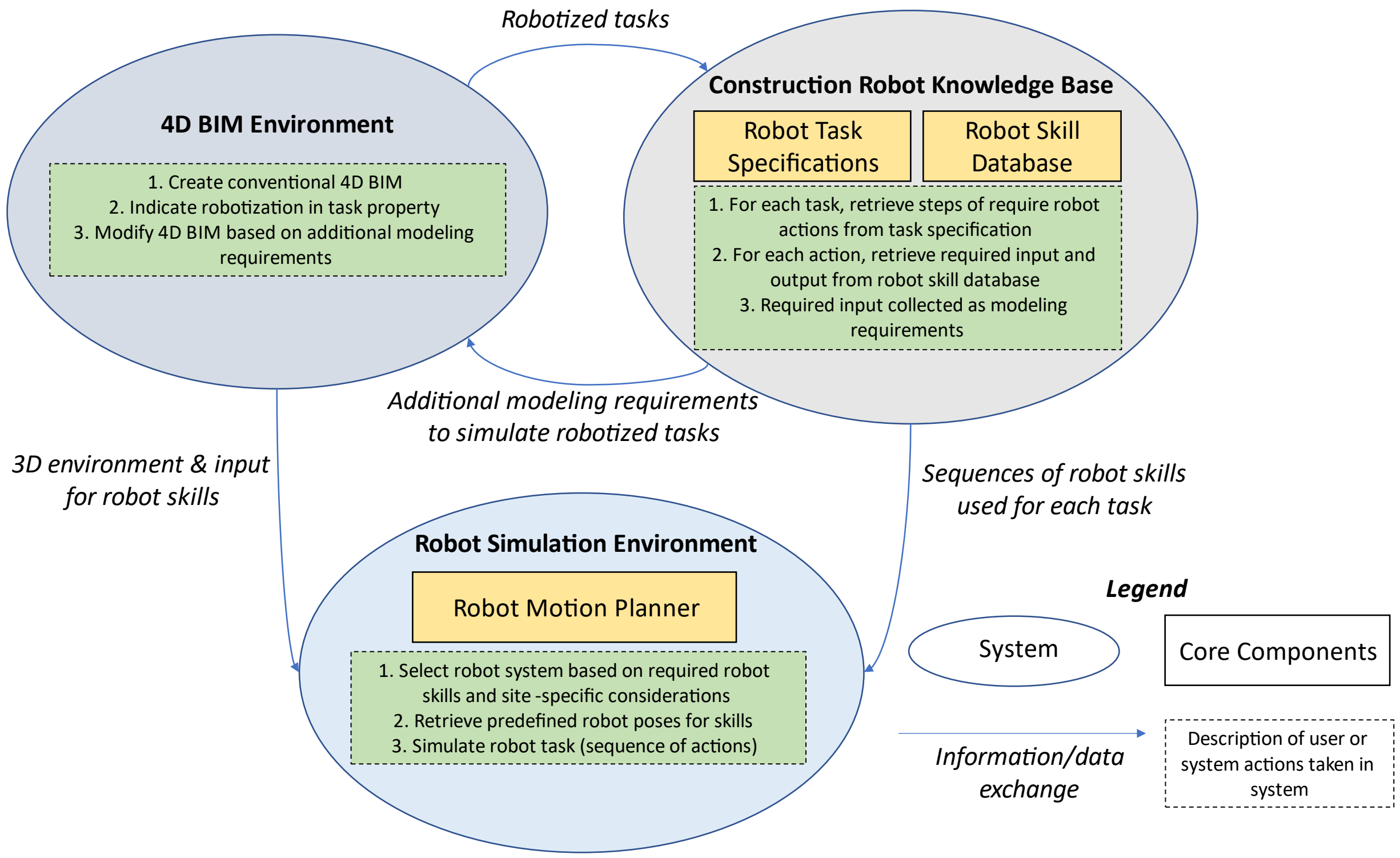

Figure 1. Framework for the integration of 4D BIM and robot task planning

Figure 1 illustrates the proposed framework that integrates 4D BIM and robot task planning via the construction robot knowledge base. This framework is composed of three major systems that produce different types of information and interact with other systems by exchanging and transforming information. **4D BIM environment** is where most of the information about the construction environment is created and high-level robot task planning is performed. Using a 4D BIM platform, such as Autodesk Navisworks [68] and Synchro [69], a 4D BIM model is created by connecting 3D building elements with tasks in the construction schedule. Then, high-level robot task planning is performed by selecting the tasks to be robotized, specifying start and finish dates, and selecting construction robot task specifications. A user will need to provide various additional inputs, such as material storage areas, areas prohibited for robot navigation, and detailed work sequences, that are not commonly specified for human workers. To link the conventional 4D BIM model with the construction robot knowledge base, this study augments "robotization" and "task specification ID" as additional properties of construction tasks (see construction schedule in Figure 2). "Robotization" is a property with a Boolean (True/False) value that indicates whether a task is to be performed by a robot, and "task specification ID" is a property with a text value that links a construction task with a task specification in the construction robot knowledge base. **Construction robot knowledge base** is the key enabler where generic knowledge about robot task performance and robot skills are stored. This knowledge base contains a robot skill database and a collection of robot task specifications. A robot task specification provides a robot-oriented instruction to perform a task, such as the sequence of actions, skills used by the actions, and site-specific or task-specific variations applied to implement the skills. At least one robot task specification with a task specification ID should be prepared, and multiple task specifications may exist for one task. The robot skill database explains what inputs and robot capacities are required to execute skills. To allow the robot to retrieve all the information required for detailed simulation, all the inputs needed to run the skills become additional 4D BIM modeling requirements. Then, the user updates the 4D BIM model based on the additional modeling requirements derived from the construction robot knowledge base. The resultant 4D BIM model is converted into a robot simulation world where a robot's detailed movements, manipulations, and sensor-based perception can be realistically simulated. Importantly, in this way, robots in simulations can retrieve sufficient information to perform tasks directly from the simulation world. **Robot simulation environment** is an ROS-compatible environment where robots as well as their work environment called a world are imported. From the construction knowledge base, the generic description of the selected skills (including names of robot skills used, input arguments for the skills, and required robot capacities) is transferred to the robot simulation environment. In the robot simulation environment, robots equipped with the required skills are selected, the robots' postures to execute the actions are predefined, and the overall task simulation is performed. Each robot interacts with objects in the work environment by extracting necessary information or physically manipulating them. Finally, the robots performing the tasks and their interactions with construction site objects can be visually analyzed and recorded for quantitative analysis of the construction plan.

### 4.2. Information flow between 4D BIM, construction robot knowledge base, and robot simulation environment

Figure 2 presents detailed information flows between the three systems with lines representing the generation and transfer of information (lines A1 to A6 and lines B1 to B5 in Figure 2). A1 to A6 present information exchanges between the 4D BIM environment and the construction robot knowledge base leading to the generation of additional 4D BIM modeling requirements and the creation of a robot simulation world based on the 4D BIM satisfying the additional modeling requirements. B1 to B5 present the information flows from 4D BIM and construction robot knowledge base into the robot simulation environment to select robots and simulate robot operations within the simulation world according to the robot task specifications.

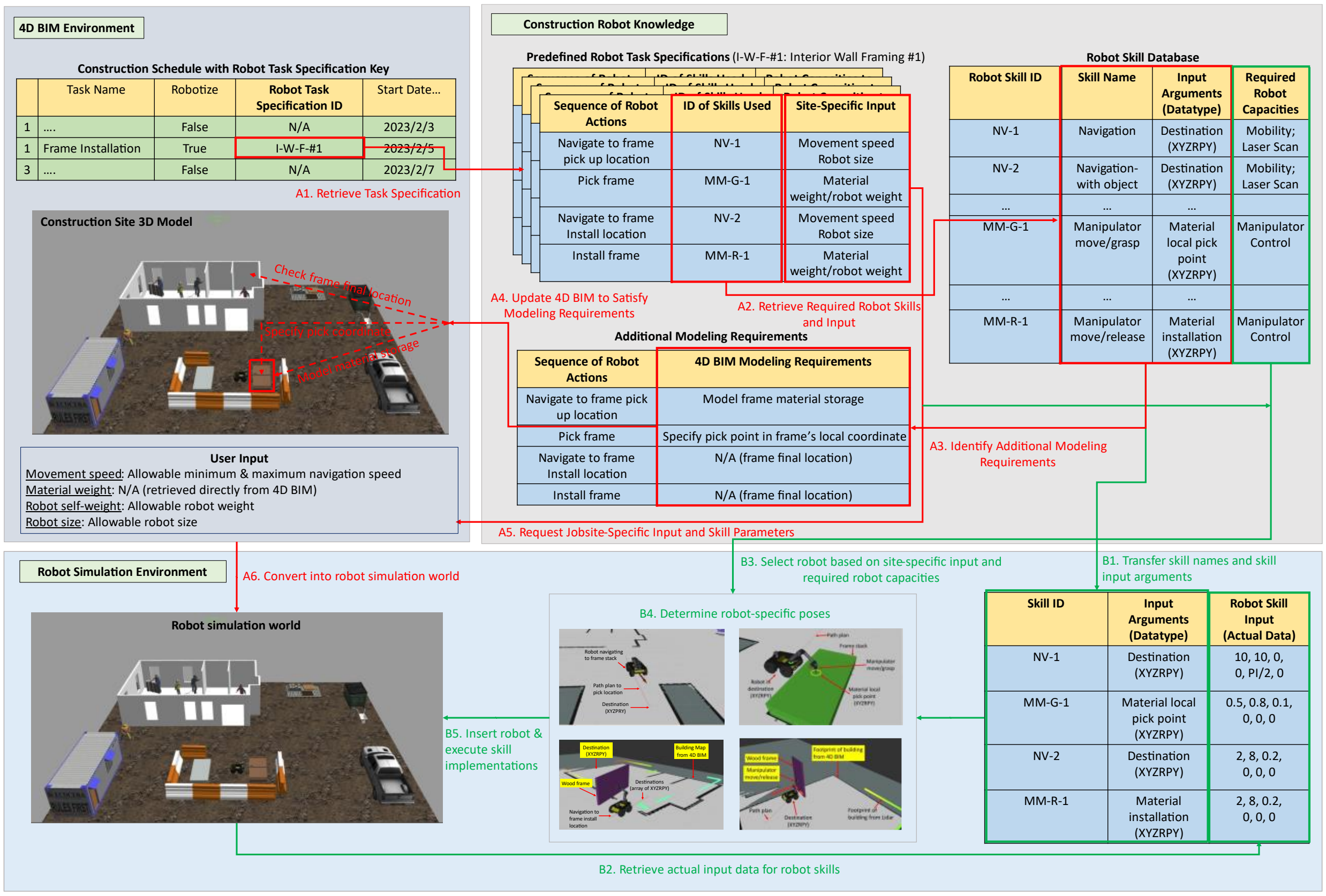

Figure 2. Information flow between three major systems

The process starts by selecting tasks to robotize from the construction schedule and ends by completing the simulation of the selected tasks in a robot simulation world. In **A1 retrieval of task specification**, a user determines tasks to be robotized and specifies robot task specification IDs for the selected tasks. In the example in Figure 2, indoor wall framing is selected for robotization and given I-W-F-#1 (Indoor-Wall-Framing-#1) as its task specification ID. Multiple specifications, such as I-W-F-#2 and I-W-F-#3, can be prepared if there are multiple methods and tools to install wall frames. The task specification ID is used as the key to find the corresponding task specification in the robot knowledge base. A robot task specification in this study contains minimal information, such as the sequence of actions, skills used, and additional site-specific or task-specific input to execute each skill, which can be extended in future studies. I-W-F-#1 describes that this task is completed by four consecutive actions “navigate to frame pick up location”, “pick frame”, “navigate to install location”, and “install frame”. These actions use different skills that are “navigation” (NV-1), “move manipulator and grasp” (MM-G-1), “navigation with object” (NV-2), and “move manipulator and release” (MM-R-1). **A2 retrieval of required robot skills and input arguments** uses the robot skill IDs in I-W-F-#1 to identify matching skills and their input arguments from the robot skill database. The minimal robot skill database in this study has skill names, skill IDs, input arguments, and required capabilities for interior wall framing installation only. These skills are executed by a robot extracting the required input data from the simulation world. Since a robot simulation world is generated from a 4D BIM model, all the inputs for the skills should be modeled in the 4D BIM model. For that, **A3 identification of additional modeling requirements** generates a list of objects or information that should exist in the 4D BIM model to execute the skills. As shown in the additional modeling requirements table in Figure 2, a frame material storage should be modeled to execute the "navigate to frame pick up location" action using NV-1 skill which needs a 3D coordinate of the destination as the input. "Pick frame" requires the user to specify a local coordinate in the frame so that a robot can grasp the material (see Figure 5). "Install frame" does not need additional modeling in the 4D BIM model since its MM-R-1 skill can retrieve the final 3D coordinate of the frame from building elements. In **A4 updating 4D BIM to satisfy modeling requirements**, a user updates the BIM model by addressing all the additional modeling requirements. For I-W-F-#1, a frame material storage was modeled together with a robot's pickup location next to the storage, and a local coordinate for pickup was specified for frame materials. Additionally, **A5 requesting jobsite-specific input and skill parameters** allows the user to determine additional restrictions (e.g., allowable robot size and weight) and tune parameters (minimum and maximum navigation speed) for robot actions considering the construction site conditions. After updating the 4D BIM model, **A6 conversion into robot simulation world** converts the 4D BIM model with additional user input into a robot simulation world. Section 4.3 describes the details of the conversion process. As a result of A1-A6, a robot simulation world representing the construction site conditions and containing all the input for robot task planning is created. The world becomes a ROS-compatible work environment for robots selected in the following steps. Following B1-B5, a detailed robot task simulation is performed. Descriptions of the skills are transferred from the previous steps (**B1**), and actual input values for the skills are extracted from the simulation world (**B2**). Based on the descriptions still highly generic, a user can select robots meeting the required robot capabilities and site-specific requirements (**B3**). For the selected robots, initial, intermediate, and final poses are determined for the skills (**B4**). Finally, the robots are inserted into the simulation world to perform the tasks (**B5**). retrieving information from the world. More details about the frame installation example are presented in section 4.4.

## 4.3. 4D BIM model conversion to robot simulation world

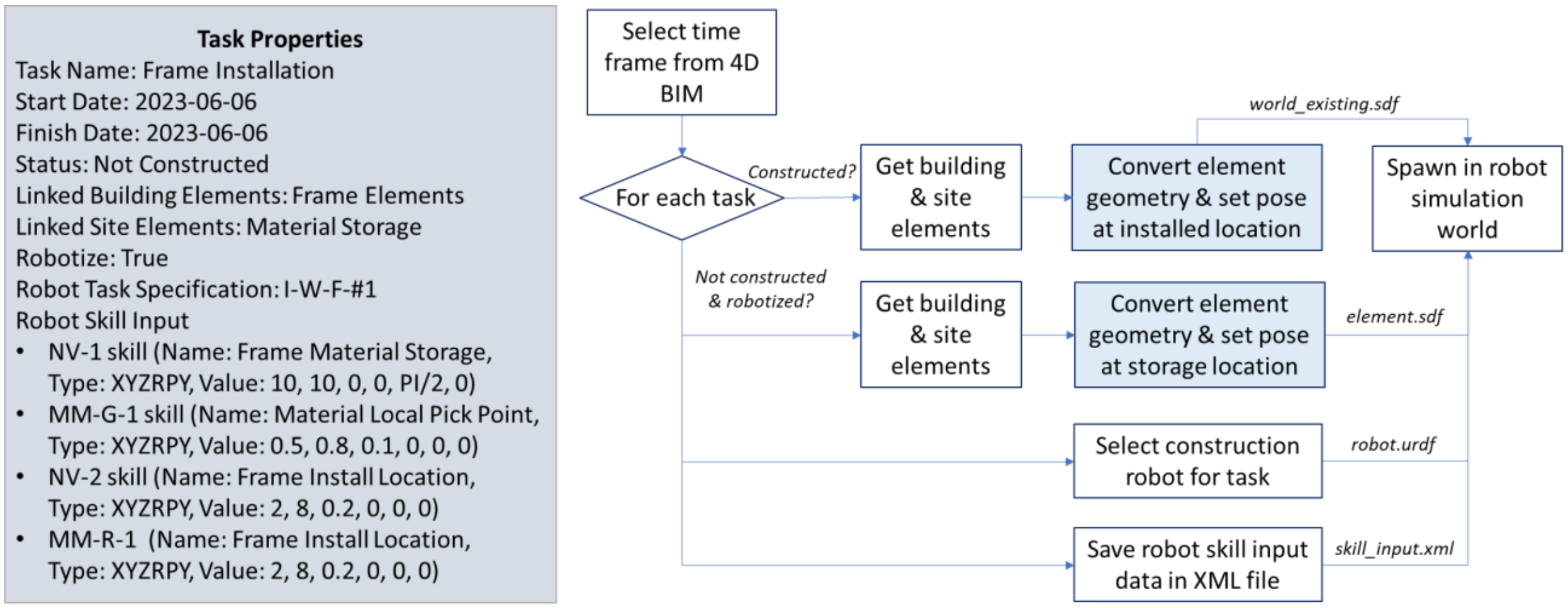

Figure 3. Task properties (left) and conversion of building and site elements (right)

Figure 3 shows the overview of the process of converting a 4D BIM into a robot simulation world. Utilizing the properties of construction tasks, this process performs sorting and conversation of building elements and construction site objects into ROS-compatible SDF files that can be spawned in a Gazebo simulation with the construction robot. Figure 3 left shows an example of properties for the frame installation task. If the status property is “constructed”, building elements and site objects attached to this task are combined into one sdf file representing the existing construction environment. Poses of these elements are set at the installed locations. If the status is “not constructed” and robotize is “true”, building elements and site objects to be constructed by the robot are converted into individual sdf files. In the case of the wall frame installation task, the poses of the frame elements to be constructed are set at the material storage locations. The spawning locations can be identical to the installed locations for tasks not moving the materials, such as painting and inspection. The conversion from BIM elements to SDF files is described in Figure 4. Other important task properties related to robot skills are stored in an XML file that can be easily parsed when a robot intends to execute the skills in the simulation world. The construction robot selected for the task is separately spawned in the simulation world. A robot model initially in Unified Robot Description Format (URDF) format is internally converted into an SDF file for inclusion in Gazebo simulation.

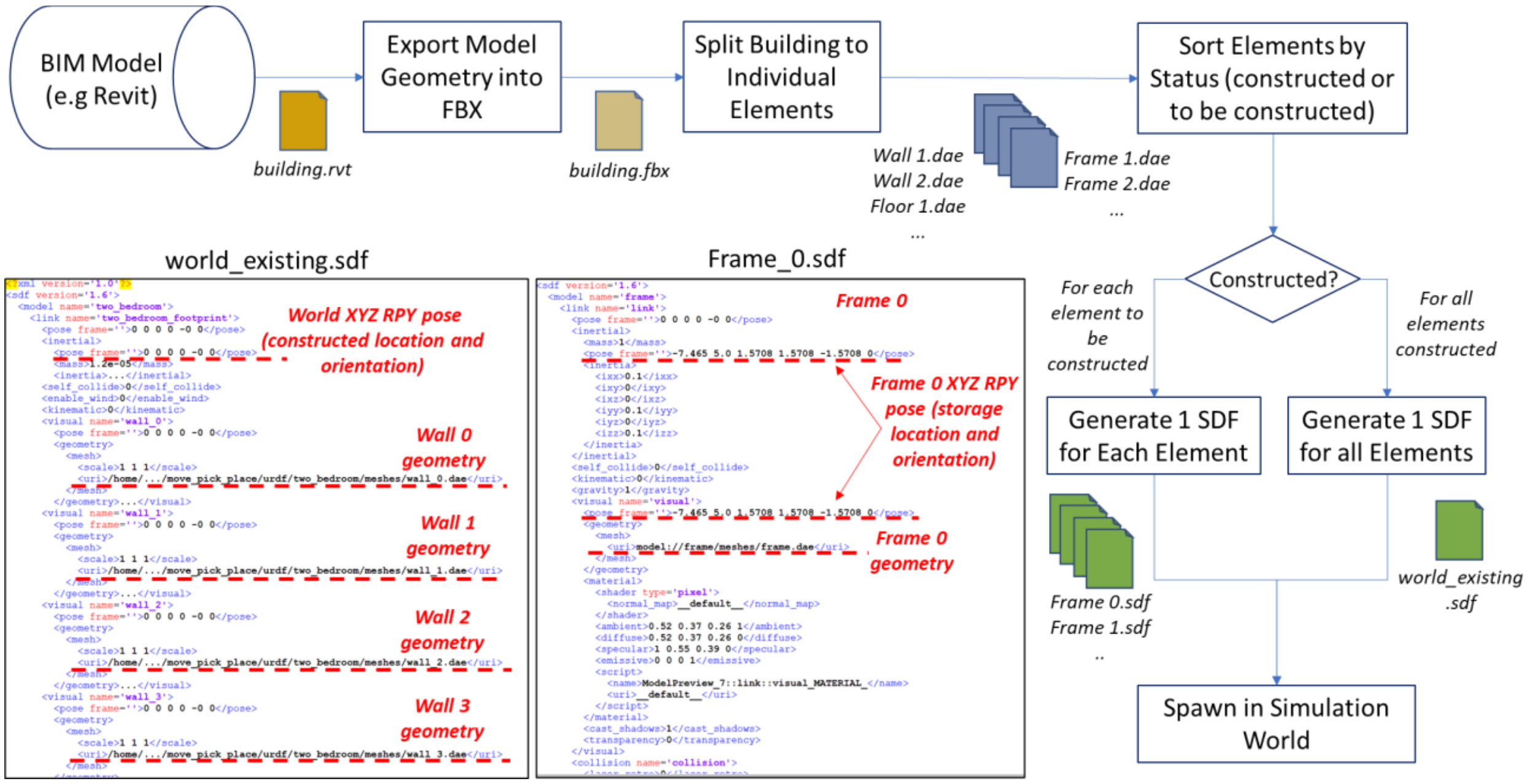

Figure 4. Building and site element conversion pipeline

Figure 4 illustrates the BIM-to-SDF conversion process with examples. The geometry model for the building elements and site objects are exported in FBX file format from a BIM application, such as Autodesk Revit. The FBX file is opened in an application supporting DAE files (e.g., Blender 3D) and they are converted to a DAE file format. The DAE file for the building entire building is subsequently split into separate DAE files of individual elements such as wall 1, wall 2, floor 1, and floor 2. For this, we created a simple Blender-based Python script to split one DAE file into individual DAE files containing geometric information, element names, and unique ID’s. The individual elements are then sorted depending on whether they are constructed or yet to be constructed elements. For constructed elements, they are parsed through another Python script we created for generating an SDF file that merges the individual DAE files to create an SDF file for the existing world as shown in Figure 4 as world_existing.sdf. The individual DAE file for each element to be constructed is parsed through the Python script to create an SDF file for each of the elements as shown in Figure 4 as frame_0.sdf.

**4.4. Execution of frame installation task with four actions**

A task is performed by completing multiple actions, and each action is performed by executing one or more skills utilizing input data retrieved from the simulation world. Like recent studies following the ROS framework [70], the implementation of robot skills in our study utilizes multiple ROS-based software packages including Gazebo [60], MoveIt motion planning framework [71], Adaptive Monte Carlo Localization (AMCL) [72], and Robot Visualizer (RViz) [73]. Gazebo is a 3D physics-enabled robot simulator that realistically mimics a real-world environment. In this study, a robot model and the construction environment are separately spawned in Gazebo simulation. The robot model with all the sensing, navigation, and manipulation capabilities can interact with objects in the construction site. MoveIt motion planning framework is a software package to plan and control robot manipulators. With the inverse kinematics library in MoveIt [71], our study determines initial, intermediate, and final robot arm poses for each action and generates collision-free motions between the poses. AMCL is a probabilistic localization package for a mobile robot. A ROS implementation of the AMCL package provides functions to estimate the current position and plan and execute autonomous navigation with an obstacle avoidance capability. Throughout the robot task planning and execution, RViz visualizes the robot's internal states, including estimated robot location and orientation, joint angles, sensor messages generated from simulated cameras and LiDARs, etc. Using these ROS packages and according to the task specification I-W-F-#1, this study focuses on performing a detailed simulation of frame installation task that comprises four major steps which are "navigate to the frame pickup location", "pick frame", "navigate to the frame install location", and "install frame". Sections 4.4.1-4.4.4 explain how these four actions are performed using different skills and how the robot skills utilize information retrieved from the simulation world. Like task specifications, this study arbitrarily named the four skills (NV-1, NV-2, MM-G-1, MM-R-1) considering the context of the frame installation task, and these names can change as more skills are added to the robot skill database. Also, detailed simulation of different skills for construction tasks may require the utilization of other ROS packages, such as object detection [74], object localization [75], and robot state monitoring and control [76]. Significant modification to these existing packages or the development of new packages would be necessary for unique and complex construction tasks.

4.4.1. Action 1: Navigate to the pickup location action with NV-1 skill

As the first action in the frame installation task, the robot navigates to the frame pickup location in the material storage area using conventional navigation with pre-determined arm poses. This action uses AMCL and MoveIt packages for navigation and arm pose control, respectively. Figure 5 shows the default pose and pose during navigation (NV-1 pose) determined using the MoveIt Setup Assistant [77] ensuring the arm does not collide with or obstruct the 2D laser scanner and camera. As the input to perform the navigation, the AMCL algorithm requires the initial pose, destination, 2D metric map, and 2D laser scan. A 2D metric map for the simulation world is generated using a PGM map creator package [78]. 2D laser scan data is generated from the simulated 2D LiDAR sensor attached to the robot. The robot's pose, including the initial pose, is continuously estimated by comparing the 2D metric map with the shape of the environment obtained from the 2D LiDAR. The frame pickup location is modeled in the 4D BIM model satisfying additional modeling requirements, and it is retrieved from the simulation world as the destination when this skill is executed. Based on these input data, the AMCL navigation package generates a navigation path and moves the robot while detecting obstacles with the 2D LiDAR and updating the navigation path. The XYZRPY format for poses explains the XYZ location and RPY orientation of the robot or other objects.

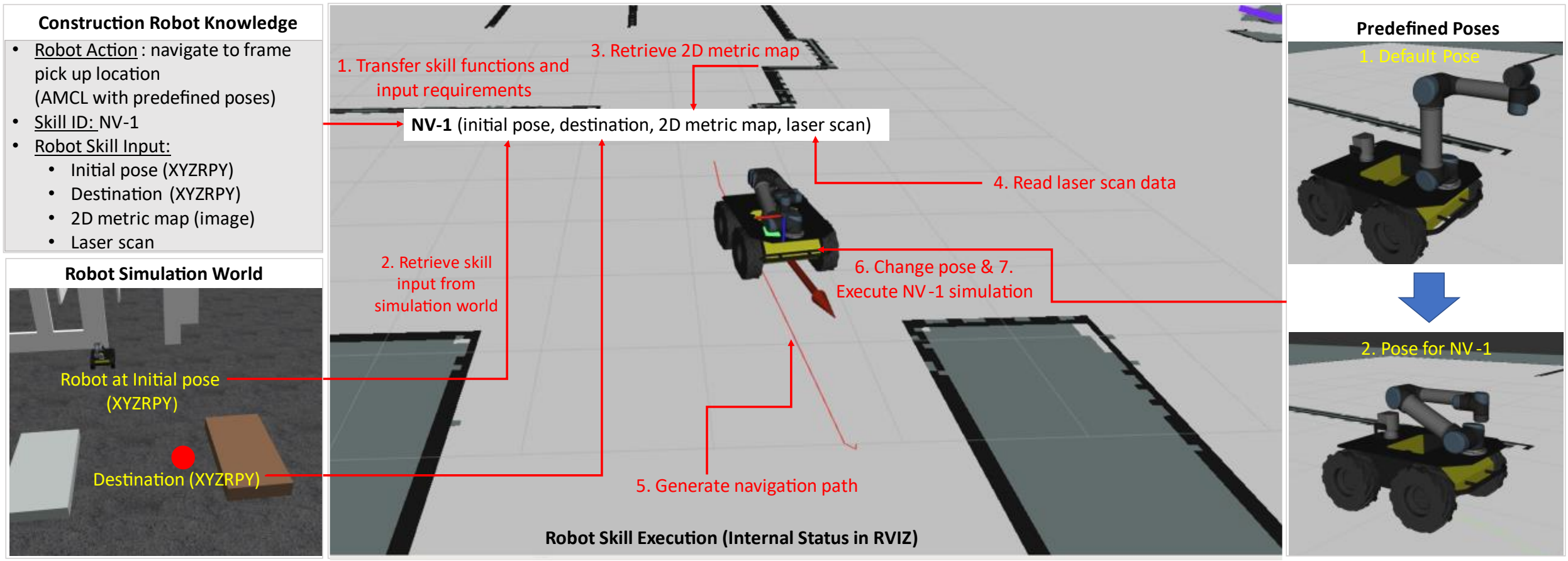

Figure 5. Robot navigating from the initial location to the pickup location with NV-1 skill

4.4.2. Action 2: Pickup frame action with MM-G-1 skill

After the arrival at the pickup location, the robot picks up a wall frame with the manipulator using the move manipulator and grasp skill (MM-G-1). This action uses MoveIt to determine arm poses and generate arm trajectories between the poses. As shown in Figure 6, the robot arrives at the destination with the arm in NV-1 pose. Then, the robot arm goes into intermediate pose 1 and intermediate pose 2 to properly position the robot arm above the material pick point. The material pick point is a local coordinate within the frame that is retrieved from the simulation world as the skill input. Then, the robot picks the frame at the material pick point. Mimicking the result of frame grasping, the frame pickup motion was achieved by simply linking the end effector of the arm with the pick point of the frame using the ROS Gazebo link attacher [79]. This simplistic approach was employed to reduce the complexity of realistically simulating the grasping arm motion.

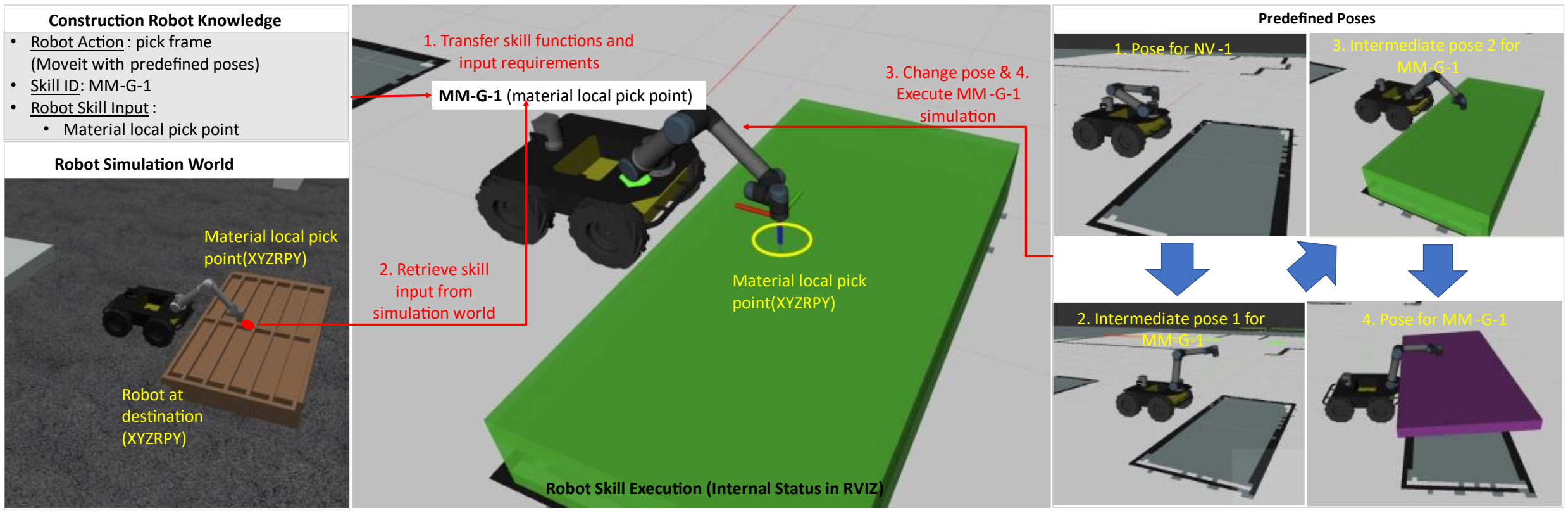

Figure 6. Robot picking frame with MM-G-1 skill

4.4.3. Action 3: Navigate to frame install location with NV-2 skill

After picking the frame, the robot carries the picked frame from the storage location to the install location. While NV-1 is a plain navigation without a material, NV-2 is a navigation carrying a material. The robot first rotates the material attached to its arm and goes from the MM-G-1 pose to the NV-2 pose as shown in Figure 7, and it navigates to the frame install location. Like NV-1, the navigation to the install position is completed using AMCL. After the robot arm goes into the NV-2 pose with the frame, the skill function is activated to receive four input arguments (initial pose, destination, 2D metric map, laser scan) for AMCL. The destination is retrieved from the simulation world as the destination input for NV-2 skill.

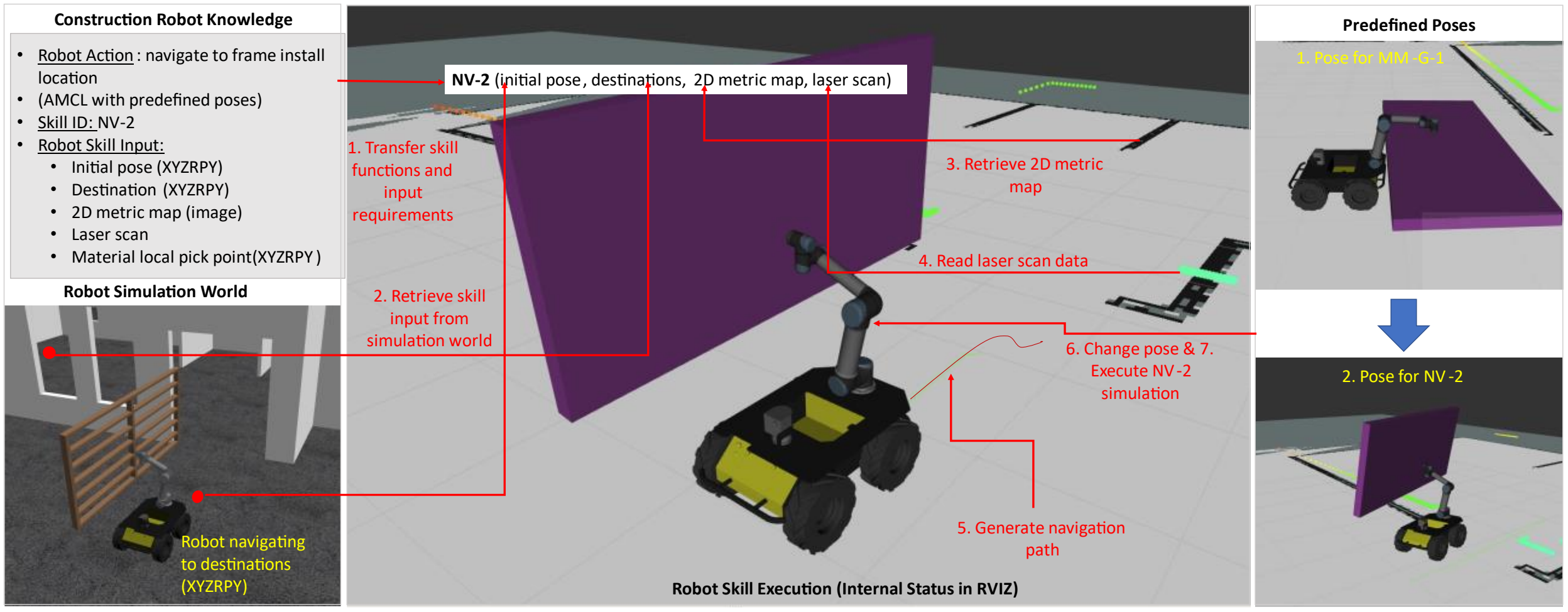

Figure 7. Robot navigating from pickup location to installation location with NV-2 skill

4.4.4. Action 4: Install frame (MM-R-1)

Finally, the install frame robot action fixes the frame at the install location using the move manipulator and release skill (MM-R-1). Similar to MM-G-1 skill, MM-R-1 relies on predefined poses shown in Figure 8. From the pick pose, the robot first goes into an intermediate pose to orientate the frame in a manner to fix it and into the MM-R-1 to place the frame at the install position. The ROS Gazebo link attacher [79] is then deactivated to detach the frame from the robot's end effector and reactivated to affix the frame at the install position.

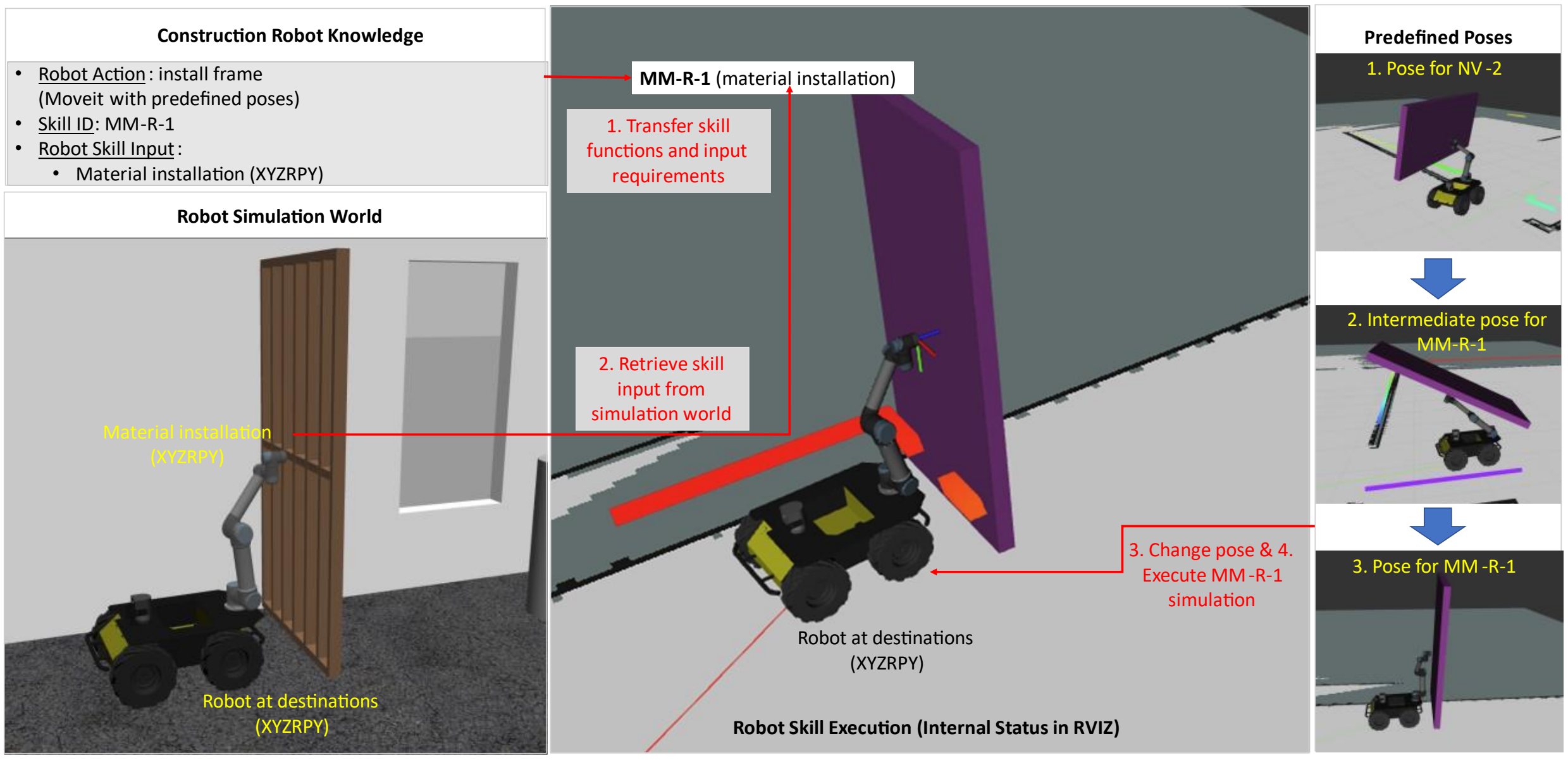

Figure 8. Robot placing the frame with MM-R-1 skill.

## 5. Case study

This section presents a case study describing the utilization of information from 4D BIM by a mobile robot to perform high-level task planning and detailed simulation of a frame installation task. The simulation for this case study was performed on a computer running Ubuntu 18 Linux setup with a Melodic version of ROS. Figure 9 shows a ROS-enabled Husky mobile robot setup and the task specification for the frame installation. The task specification describes the actions to be executed by the Husky robot with the skills and input data required to complete the frame installation task. The Husky is a medium-sized four-wheeled mobile robot equipped with an LMS-111 2D Lidar scanner, an Intel RealSense depth camera D455, and a Universal Robot 5 (UR5) manipulator. The Husky's wheels and ability to navigate rugged construction site terrain make it well-suited for our case study. The 2D Lidar scanner provides essential scan data to support Husky's localization and navigation planning. The RealSense camera captures RGB and RGBD data, facilitating tasks like object detection, localization, and improving robot navigation. The UR5 is a ROS-enabled robotic arm with six rotating joints capable of executing intricate maneuvers for picking and installing the frame. Since the Husky robot and its components are ROS-compatible, we pre-defined frame installation task instructions as retrievable ROS actions. With substantial effort, any ROS-compatible robot can be adapted for similar or different tasks. Details on robot actions, skills, and data input have been provided in section 4.

Figure 10 shows the Husky robot alongside two carpentry work crews performing roof installation and the preparation of wood frames for interior wall partitioning. The carpenters 1 and 2 move around the construction environment to get the materials required for their various tasks. The site supervisor also makes routine movements around the construction site to monitor the ongoing activities. The Husky robot has been assigned the task of picking the frame material from the storage location and installing them in the building.

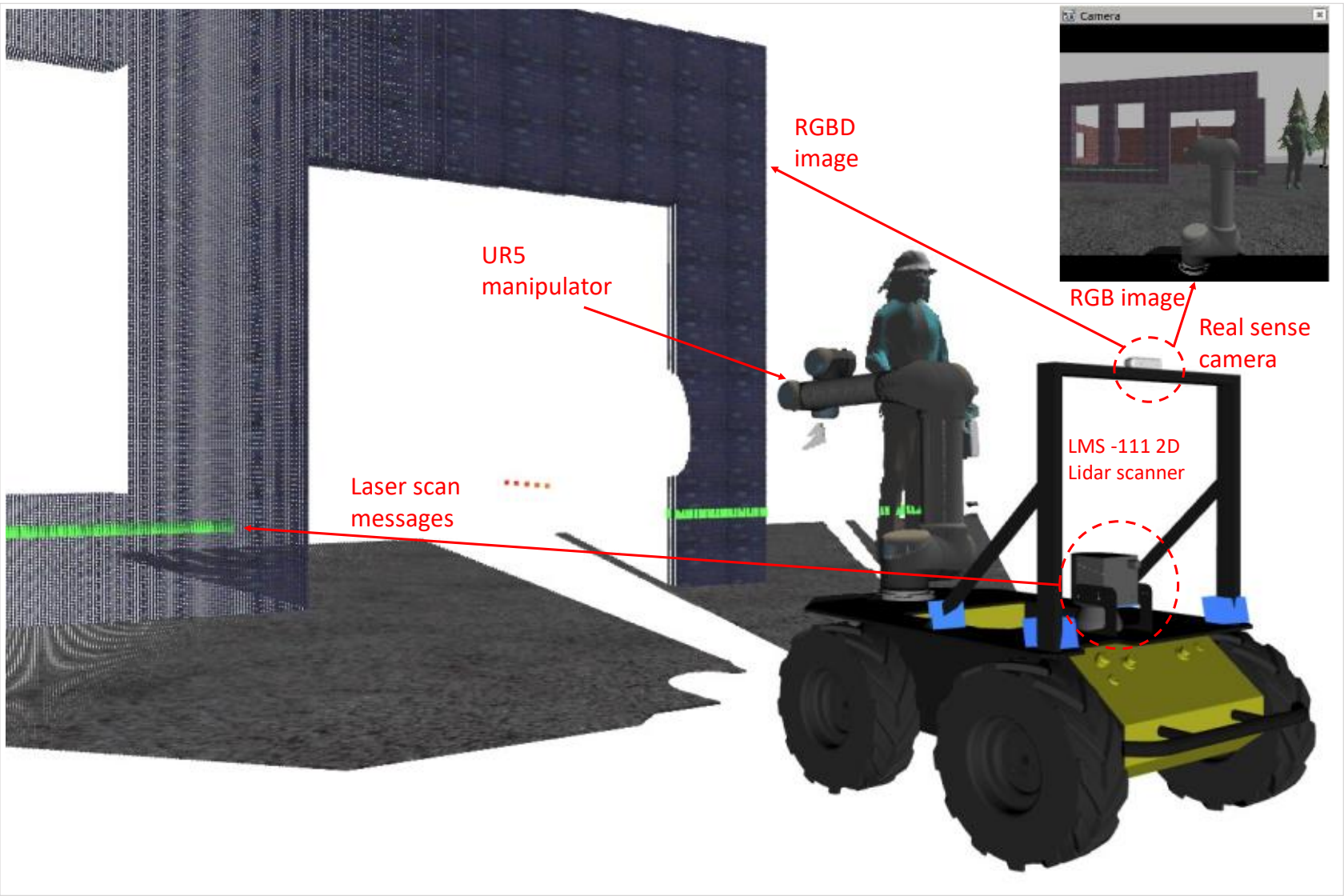

Robot task specification

| | Robot actions | Robot skills needed | Robot skill input data (xml) |
|---|---|---|---|
| 1 | Navigate to frame pick up location | Navigation (NV-1) | [robot initial pose, destination, 2D metric map, laser scan] |
| 2 | Pick frame | Manipulator move/grasp (MM-G-1) | [material local pick point] |
| 3 | Navigate to frame install location | Navigation (NV-2) | [robot initial pose, destinations, 2D metric map, laser scan] |
| 4 | Install frame | Manipulator move/release (MM-R-1) | [material installation] |

Robot skill Input data (xml)

Robot initial pose[0.0, 0.0, 0.0, 0.0, 0.0, 0.0]
Pick coordinates [2.0, 1.0, 1.0, 0.0, 0.0, 0.0]
Destination [-2.465, 3.21, 1.2, 1.57, 1.57, 0.0]
Movement speed: 1.3 m/s
Material installation_0 [-4.465, 0.20124, 1.2, 1.57, 1.57, 0.0]
Material installation_1 [-3.255, 0.20124, 1.2, 1.57, 1.57, 0.0]
Material installation _2 [-2.046, 0.20124, 1.2, 1.57, 1.57, 0.0]
Material installation _3 [-0.838, 0.20124, 1.2, 1.57, 1.57, 0.0]
Material installation _4 [0.0, -2.997, 1.2, 1.57, 1.57, 0.0]
Material installation _5 [0.0, -1.787, 1.2, 1.57, 1.57, 0.0]
...

Figure 9. Husky mobile robot setup and task specification for the case study

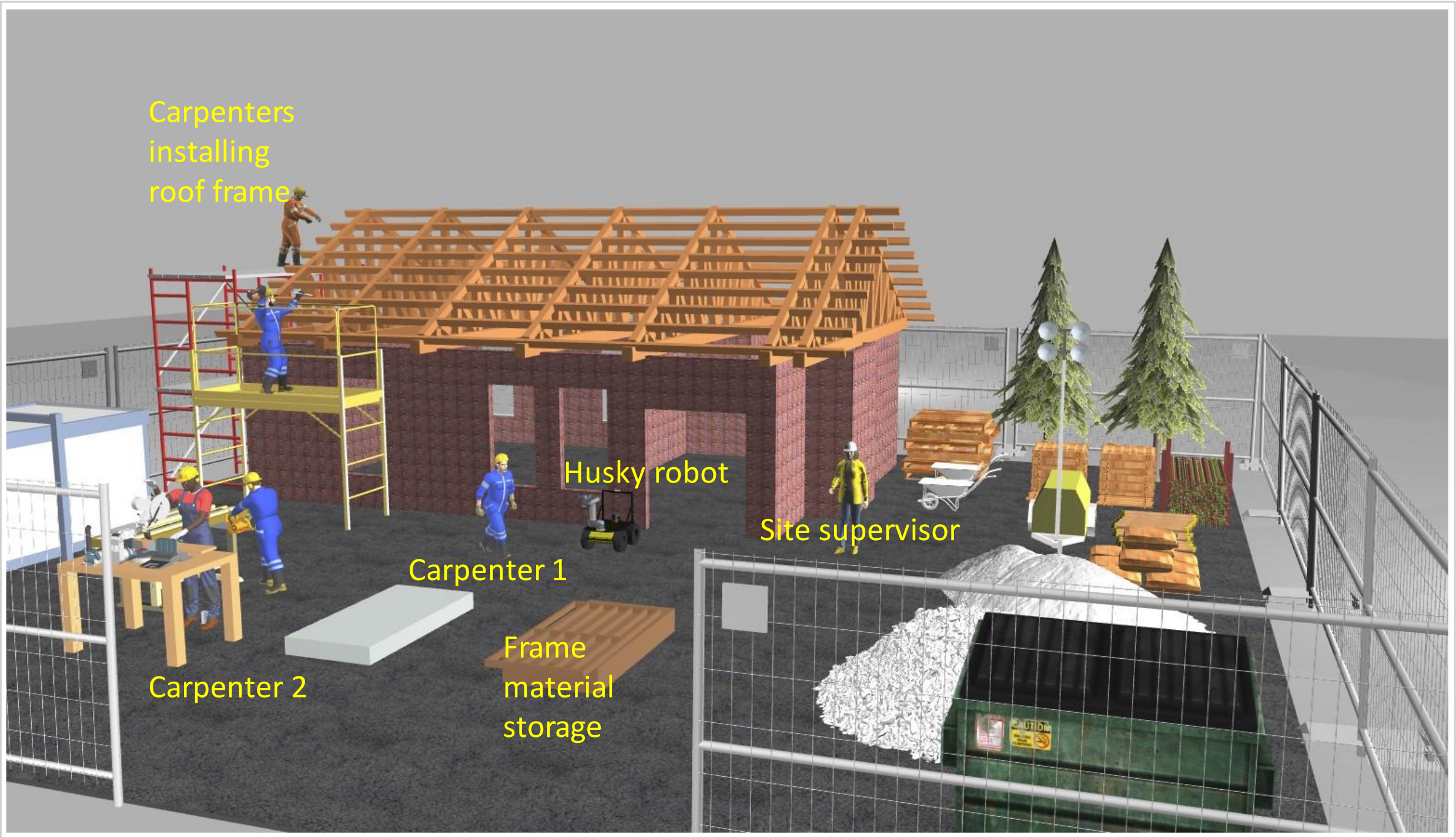

Figure 10. Case study construction environment with workers performing various tasks.

Figure 11 shows the case study construction site layout for the two-bedroom apartment, the Husky robot, the predefined robot actions to install the wood frames, and the partial project schedule showing the tasks to be performed. The construction site layout and project schedule represent the 4D model created using Autodesk Revit and Navisworks applications and converted into the robot simulation world. The construction site layout shows the general site arrangement that realistically represents a real-world situation with designated areas for material storage, equipment, and site offices. The process of modeling and creating the simulation world has been previously described in section 4.3. The partial project schedule depicts the frame interior wood partition as the next task to be performed between the 10$^{th}$ – 23$^{rd}$ of May 2022. As shown in the partial schedule, the interior wall framing task has been planned to be performed by a robot and given the additional properties "robotized" and "I-W-F-#1" in line with our methodology explained in section 4. The interior wall partitioning task is completed when 11 wood frames have been installed. The installation positions of the wood frames are marked on the layout with yellow rectangular bars. The four robot actions required to complete the task by the mobile robot are also shown in Figure 11.

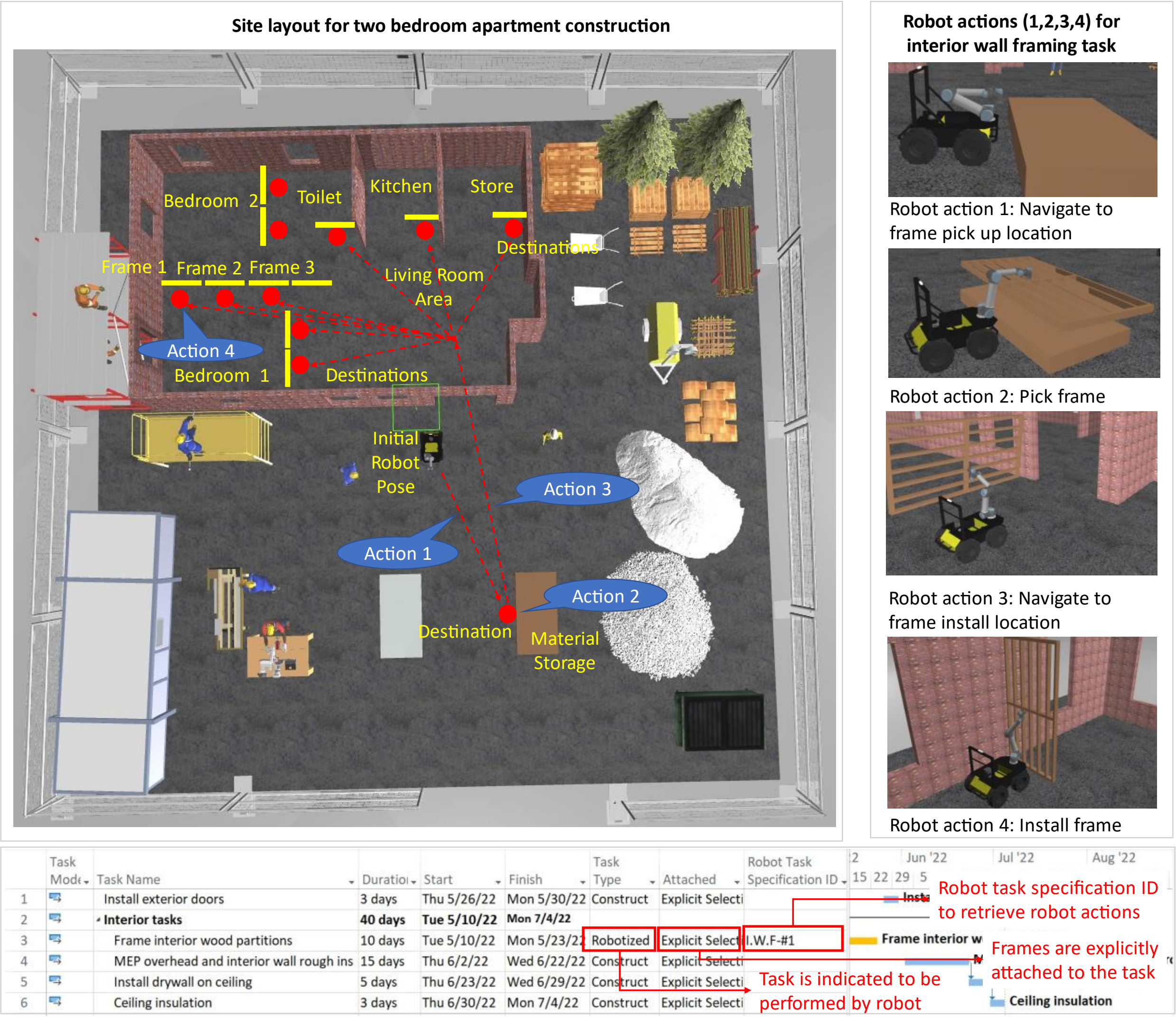

| | Task Mode | Task Name | Duration | Start | Finish | Task Type | Attached | Robot Task Specification ID |
|---|---|---|---|---|---|---|---|---|
| 1 | | Install exterior doors | 3 days | Thu 5/26/22 | Mon 5/30/22 | Construct | Explicit Selecti | |
| 2 | | **Interior tasks** | **40 days** | **Tue 5/10/22** | **Mon 7/4/22** | | | |
| 3 | | Frame interior wood partitions | 10 days | Tue 5/10/22 | Mon 5/23/22 | Robotized | Explicit Select | I.W.F-#1 |
| 4 | | MEP overhead and interior wall rough ins | 15 days | Thu 6/2/22 | Wed 6/22/22 | Construct | Explicit Selecti | |
| 5 | | Install drywall on ceiling | 5 days | Thu 6/23/22 | Wed 6/29/22 | Construct | Explicit Selecti | |
| 6 | | Ceiling insulation | 3 days | Thu 6/30/22 | Mon 7/4/22 | Construct | Explicit Selecti | |

Figure 11. Site layout, partial construction schedule, and planned robot actions to install a frame in the two-bedroom apartment.

As described in section 4, the successful installation of wood frames at designated locations requires the execution of four distinct robot actions. The frame interior wood partition task for the building is achieved through the execution of these predefined robot actions, a process that continues until all the frames have been appropriately installed. To commence the task, the robot was issued a command instructing it to perform the prescribed actions for the installation of 11 wood frames at locations marked with yellow bars in Figure 11. Frame 1 (see Figure 11) was first selected for installation to begin the interior wall partitioning task. Figure 12 shows the robot's reality versus how the robot perceives the environment using its sensors. It describes how the robot executes Action 1 by moving from its initial location to the designated pickup location next to frame material storage. During this process, the objects in the construction environment are seen by the robot's camera and reflected in their exact location on the site map. The robot uses the knowledge of object locations for obstacle avoidance and path planning. The process of creating the site map has been mentioned in section 4. On arriving at the pickup location, Action 1 is considered complete, thus commencing Action 2.

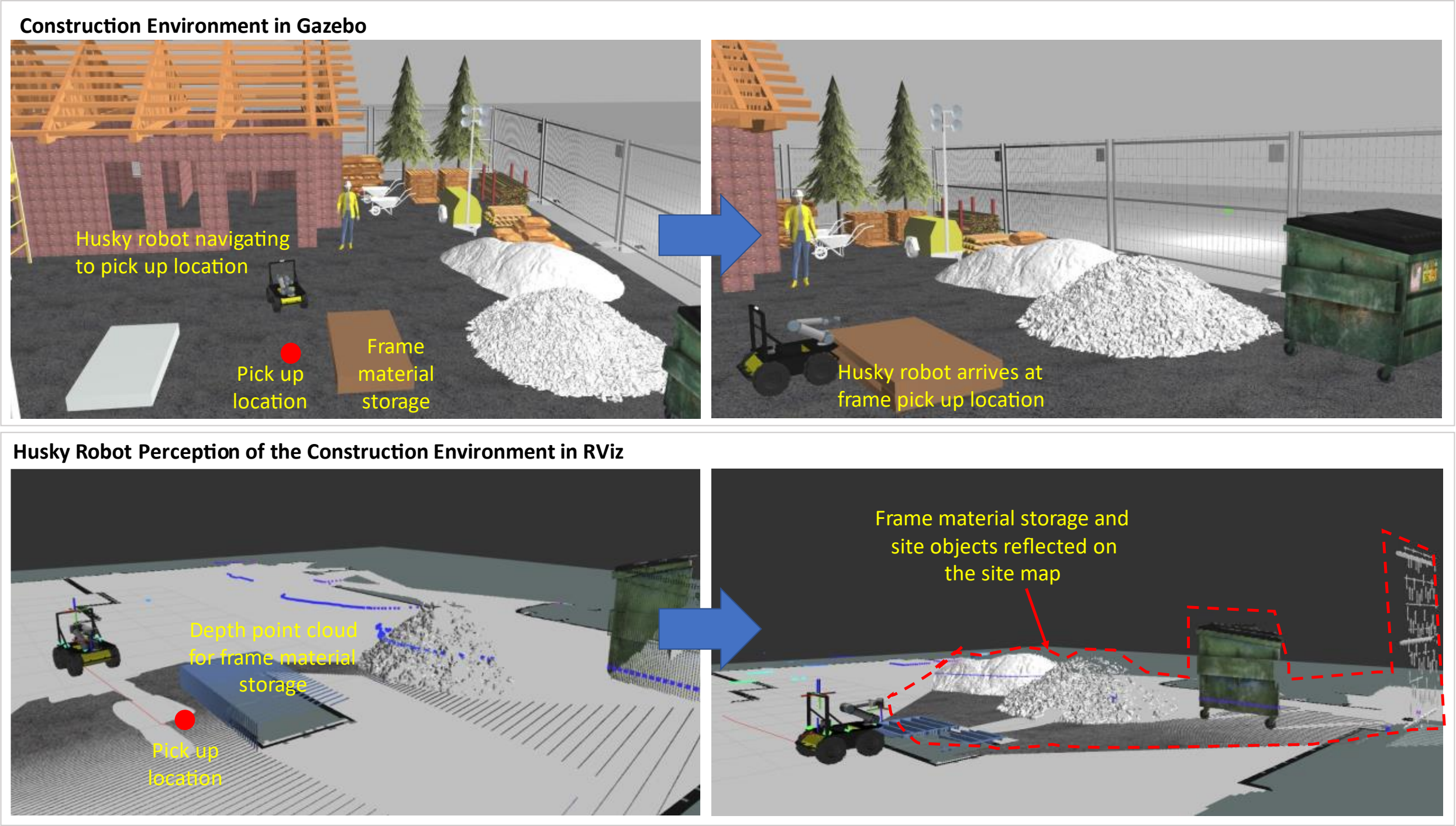

Figure 12. Robot performs Action 1 to arrive at the material pick-up location.

Figure 13 shows the robot performing Action 2, wherein it uses its robotic arm to grasp and lift the frame 1 material from the material storage. The robot registers the frame pick operation as successful when it senses the frame attached to its arm and lifts it off the material storage as shown in the image. After picking frame 1, the robot proceeds to execute Action 3 by navigating through the construction site into the building carrying the picked frame 1. When the robot gets to the building entrance and realizes that it cannot pass through the doorway with the Frame 1 material carried in its current horizontal orientation, it rotates the frame 1 material to carry it on its side to pass through the doorway as shown in Figure 14. Upon entry into the building, the robot again executes a rotation maneuver to reposition Frame 1 over its head in preparation for the installation process. The robot then continues its navigation towards the installation location for frame 1. When the robot reaches the install location, it performs Action 4 to install frame 1 as shown in Figure 15. The sequence of robot Actions 1 through 4 was repeated for varying installation locations to complete the interior building wall partitioning with the 11 frames materials.

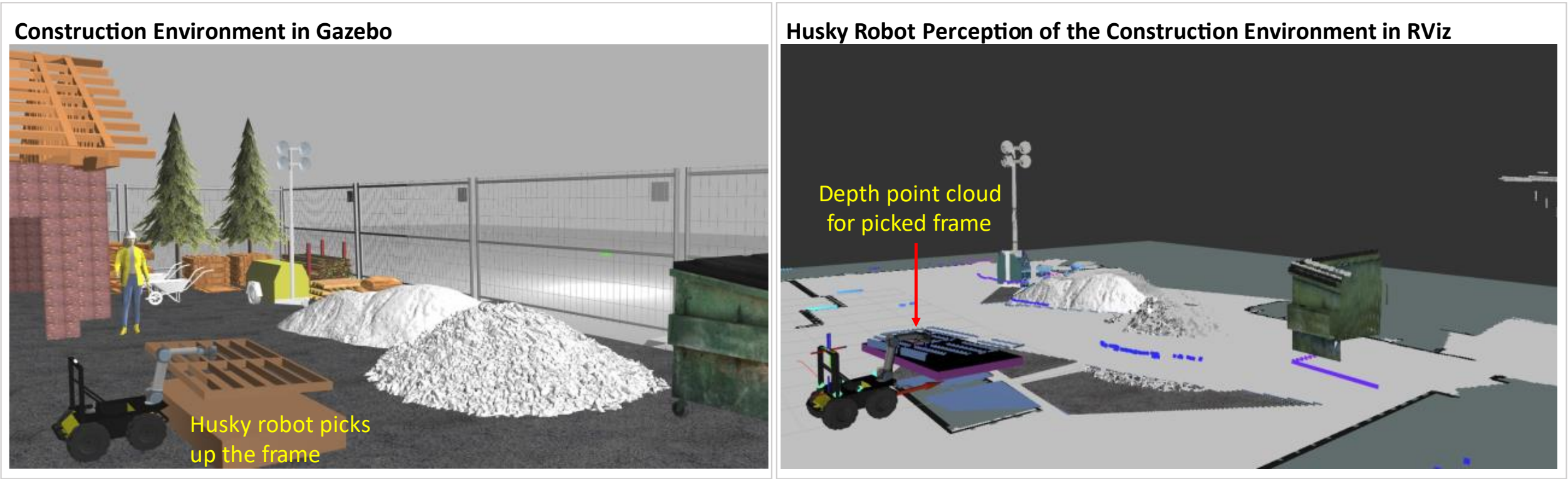

Figure 13. Robot performs action 2 to pick up frame material.

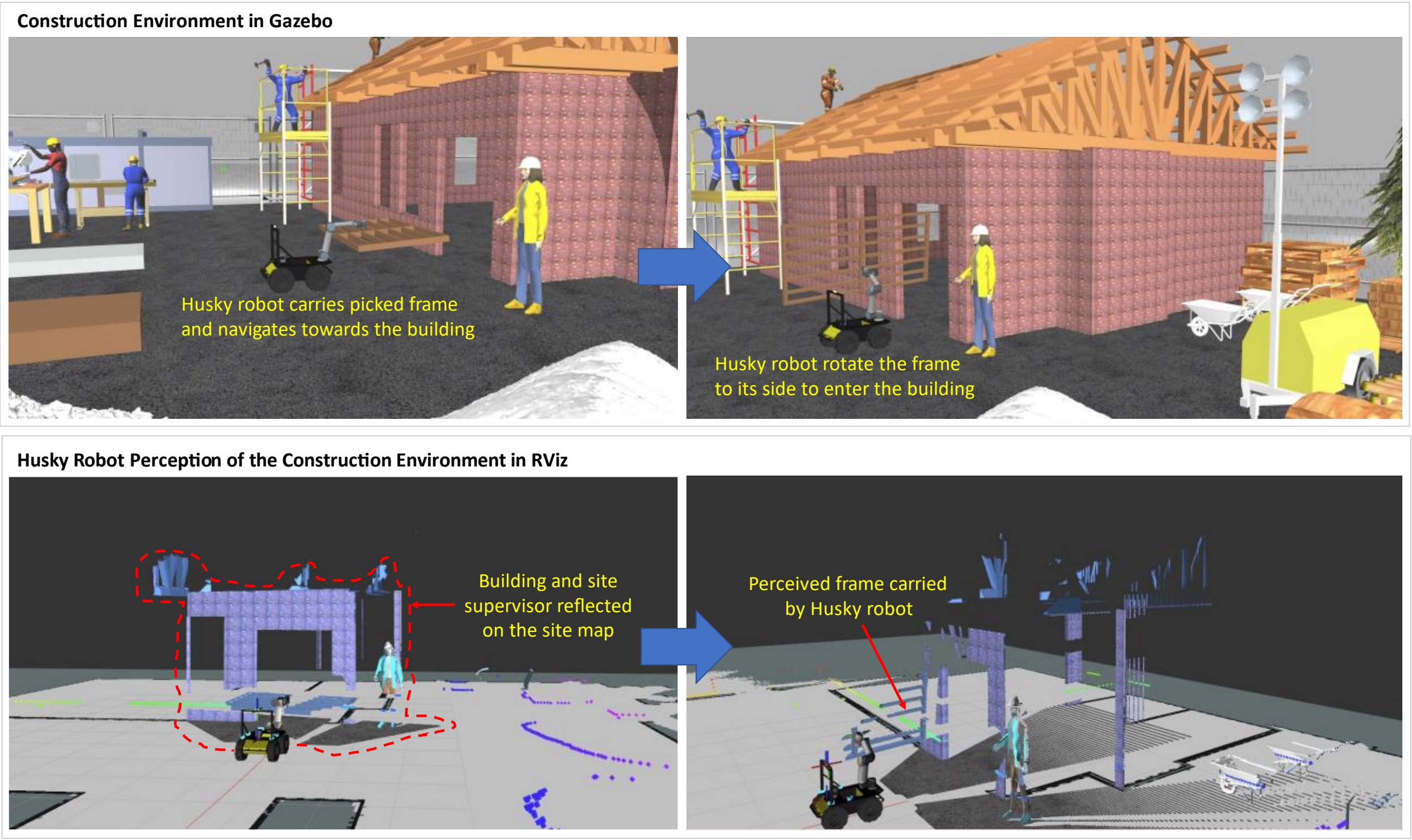

Figure 14. Robot performs Action 3 to arrive at the frame install location.

The simulation of Actions 1 through 4 to install frame 1 material took 4.48 minutes to complete. Consequently, a total of 60 minutes was expended in simulating the robot installing 11 frames to complete the interior wall partitioning. During the simulation process, we noticed the proximity between the robot and construction workers at some points. To confirm our observation, we plotted the trajectories of the robot and construction workers in the execution of their various tasks during the installation of frame 1 for more examination as shown in case one of Figure 16. We also plotted the difference in their distance apart to further analyze the simulation. This analysis unveiled instances during the simulation where the construction workers, specifically carpenters 1 and 2, were within less than 1 meter from the robot. Such proximity exposes carpenters 1 and 2 to potential incidents including collisions and struck by accidents. In response to this safety concern, various mitigation strategies were contemplated, including the adjustment of the project schedule to allocate separate time intervals for the execution of robot tasks and the separation of human work zones from robot work zones.

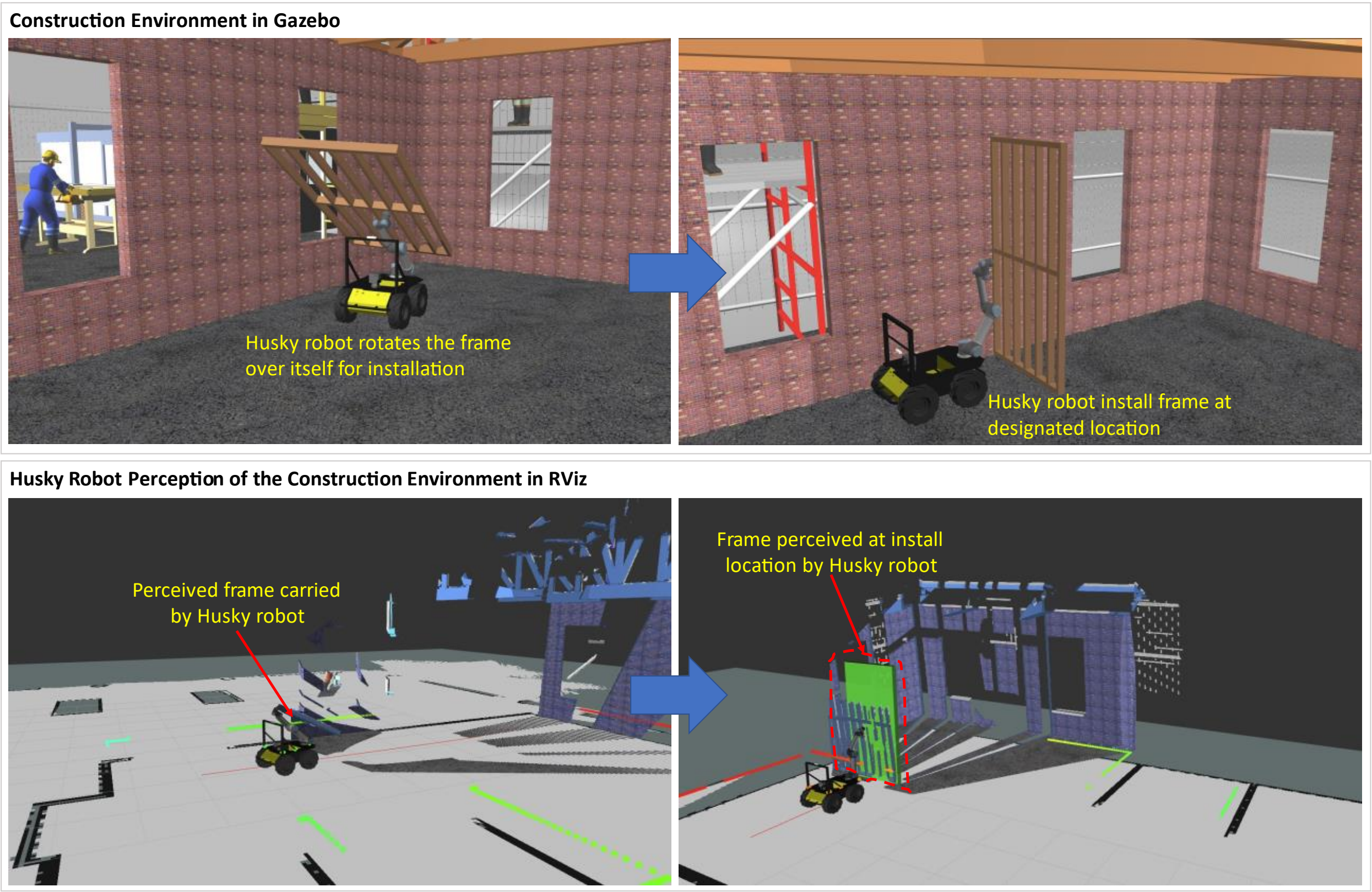

Figure 15. Robot performs Action 4 to install the frame.

However, we decided on an immediate solution. This was to relocate the frame material storage into the interior of the building to separate the robot from the human work area. Subsequently, we conducted a follow-up simulation after relocating the material storage as shown in case two of Figure 16. We observed the trajectories of the robot and construction workers during the installation of frame 1. The resulting analysis revealed that the robot consistently maintained an approximate separation distance of 3 meters from the nearest human worker, as depicted in case two of Figure 16. This strategic modification effectively addressed the safety concerns that arose during the initial simulation, mitigating the risk of inadvertent interactions between the robot and construction workers. With the material storage moved into the building, the robot took approximately 4 minutes to install frame 1.

The outcome of this case study demonstrates the significance of proper site planning for robot operations considering the spatiotemporal conditions of the construction site. It shows how planning robot operations as part of the construction site can potentially eliminate safety hazards and reduce time spent on tasks. This can lead to an increase in productivity, efficiency, and safety on construction projects incorporating robots for various tasks. This simulation met the goal of this research, which is to integrate 4D BIM and robotics to plan robot activities in performing various construction tasks.

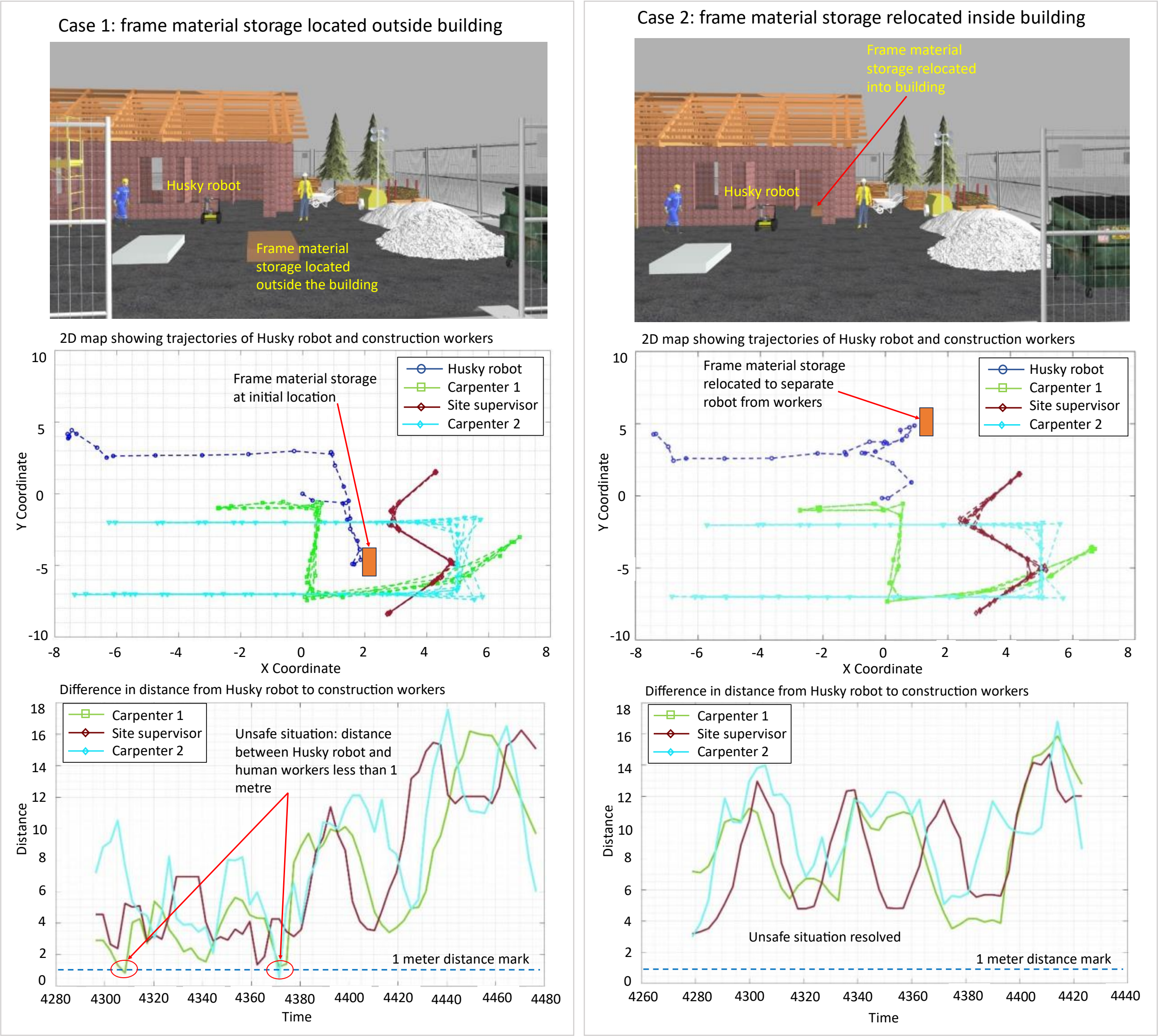

Figure 16. Cases 1 and 2 of the Husky robot and construction workers trajectories in the execution of various tasks

## 6. Conclusion and discussion

Deploying robots, especially robots with autonomous behaviors, in construction projects poses unique challenges. Unlike relatively unchanging work environments (e.g., coffee making, cooking), manufacturing construction sites experience continuous changes with structures under construction, unplanned movements of workers, multiple concurrent tasks, and construction site objects like material stacks, equipment, and temporary structures. In such dynamically changing work environments, even the same construction task may need to choose different robot configurations, control the robots differently, or cannot use any robot in limited spaces. With the current separation between robotics and 4D BIM technologies, a construction planner considering the use of robots should determine highly precise robotic movements and motions of robots based on the imagination of future construction site conditions, which is almost impossible when it is done manually.

The long-term goal of this study is to create a new process and a software platform to establish robotic plans to perform various construction tasks, simulate and evaluate robot task performances in terms of various criteria, and further design safe and productive robot behaviors adapting to all possible situations. The study presented in this paper achieves the initial step toward such construction robot task planning via systematic integration of 4D BIM and robotics. As the main contribution, this study created a new framework that describes how robot task planning can be integrated into 4D BIM-based construction planning. The systematic integration with information flows explains how robot task planning and simulation can be performed utilizing construction-related information contained in a 4D BIM model. Uniquely, this study proposed the construction robot knowledge base as the key enabler for the 4D BIM-robotics integration. The knowledge base contains task-related instructions for robots and helps derive additional modeling requirements for 4D BIM used for robot task planning. With a focus on interior wall frame installation, a case study showed that the systematic integration of 4D BIM and robotics allows the robot task planning and simulation based on the information about the construction site conditions acquired from a 4D BIM model.

There are several limitations in this study that require future studies. 1) The construction robot task specification created for a frame installation task outlines only a sequence of actions and robot skills used. For actual deployment, this overly simplified task specification should be extended to generate more complicated and practical robot behaviors. The future robot task specification can include various other capabilities, such as sensor-based perceptions, conditional behaviors, safe robot control strategies, and interaction with human workers. To define complex robot behaviors combining these capabilities, more scalable approaches, such as formal specification [76] and ontology [80,81], can be used to develop construction robot task specifications. 2) Also, after achieving the systematic integration, this study did not comprehensively explore how the developed 4D BIM-based robot task planning and simulation can be used during the construction planning process. A future study can investigate how various areas of construction planning, such as construction schedule, job site coordination, and safety planning, can utilize robotic simulation considering the overall safety, cost, and duration. 3) The simulation in this study does not perfectly mimic the real situation even using the physics-based simulator. A future study should conduct experiments with a real robot and frame materials to fine-tune the simulation.